# Two Procedures for Compiling Influence Diagrams

**Paul E. Lehner**
Systems Engineering Dept.
George Mason University
Fairfax, VA   22030

**Azar Sadigh**
Systems Engineering Dept.
George Mason University
Fairfax, VA     22030

## ABSTRACT

Two algorithms are presented for "compiling" influence diagrams into a set of simple decision rules. These decision rules define simple-to-execute, complete, consistent, and near-optimal decision procedures. These compilation algorithms can be used to derive decision procedures for human teams solving time constrained decision problems.

## 1.0 INTRODUCTION

Within some research communities (e.g., artificial intelligence), there is a growing recognition that Bayesian decision theory provides a powerful foundation upon which to develop automated and partially automated reasoning systems. Decision theory provides a compelling semantics for inference and action under uncertainty, as well as a framework for evaluating the adequacy of heuristic methods. The use of Bayesian techniques is now common place in a number of research areas, including inference (e.g., Pearl, 1987), planning (e.g., Dean and Wellman, 1991), and learning (e.g., Paass, 1991).

Most applications of Bayesian techniques involve problems that require repeated processing of similar cases. A typical example is medical diagnosis, where the objective is to use symptoms and test results to select the affliction affecting the patient. In such problems, a *probability network* is constructed that encodes a joint probability distribution over a *preselected* set of conclusions (affliction), evidence items (symptoms and diagnostic tests), and intermediate hypotheses. If decisions and outcome values are included, then the network becomes a *decision network* (a.k.a. influence diagram) that includes the possible decisions and the utility of each decision conditioned on a subset of the evidence items and hypotheses. Each decision network defines a *domain model* that identifies the possible states that could occur and the possible decisions that could be made. Constructing a good decision network involves a substantial knowledge engineering effort. However, once developed, the decision network can be used repeatedly to address decision problems which differ in the pattern of evidence that is observed.

Many decision tasks, particularly in the command and control ($C^2$) arena, involve repeated processing of similar cases (sensor interpretation, object identification, object location, IFF, etc.). These tasks are good candidates for the application of decision networks. However, many of these tasks are severely time constrained. Therefore procedures for rapid processing of decision networks, as well as rapid processing of their outputs by users, must be developed before the potential usefulness of decision networks can be fully realized.

Unfortunately, as shown by Cooper (1990), the computational complexity of computing posterior probabilities in a probability network is NP-Hard. This implies that the computational effort involved to select optimal decisions using a decision network is at least exponential with the size of the network. Consequently, in order to realistically apply Bayesian decision theory to time constrained problems, computationally simpler procedures must be developed that approximate exact Bayesian reasoning. Furthermore, many team decision tasks involve life-and-death decisions. Many believe that it is unwise to delegate such decision to machines. This implies that humanly-executable decision procedures are needed, even when normative Bayesian procedures are available.

Previous researchers have explored three distinct approaches to approximate rapid processing of decision networks. The first is to use simulation algorithms that generate approximate solutions in polynomial time (e.g., Henrion, 1988). The second (e.g., Dean and Wellman, 1991) is to partition the reasoning problem into a series of incremental reasoning steps, and to estimate the computational burden involved before each step is executed. If there is insufficient time to execute the next step, the current solution is offered as an approximation. Unfortunately, both of these approaches generate decision procedures that are not executable by people, or easy for people to understand.

A third approach is to "compile" a decision network into a set of simple decision procedures and to apply those decision procedures at execution time. This approach can be used to generate decision procedures that people can execute. Heckerman, et.al. (1989) addresses the problem of compiling decision networks by generating decision rules that specify, for each evidence state, the maximum expected utility decision. One of compilation approaches proposed by Heckerman, et.al., is to develop a *situation-action tree* in which evidence items are sequentially examined until sufficient evidence is accumulated to warrant a decision. In this paper, we develop examine two procedures for generating situation-action trees, and characterize their optimality and computational complexity properties.

## 2.0 COMPILING DECISION NETWORKS

We begin by defining some basic terms. Figure 1 depicts a small decision network. The circle nodes are *chance nodes*. Each chance node identifies a set of mutually exclusive and exhaustive propositions. The arcs between chance nodes identify the conditional probability statements that must be contained in each node. For instance, the node Typ contains the unconditional probability distribution P(Typ). The node A contains the conditional probability distribution P(A|Typ). The rectangular node is a *decision node*. Each decision node identifies an exhaustive set of mutually exclusive decisions. Arcs going from a chance node to a decision node are information arcs. They identify information that will be available when the decision must be made. The diamond node is a *value node*. A value node assigns a utility to each row in the cross product of the propositions/decisions of its parent nodes. For instance, the node Val in Figure 1 assigns a utility for each decision and state in Typ.

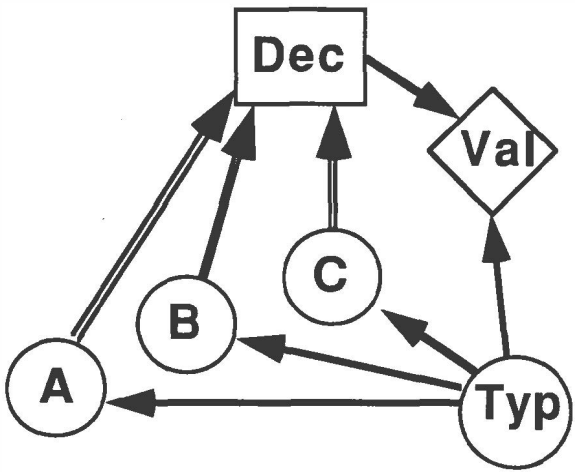

Figure 1: A Decision Network

In this paper, chance nodes that have information arcs going to a decision node are referred to as *evidence items*. An evidence item is *instantiated* when the value of that evidence item is known. For instance, A=a1 asserts that the *evidence value* for evidence item A is a1. A set of evidence values for all evidence items is referred to as an *evidence state*. For instance, the vector <a1,b2,c3> describes the evidence state where A=a1, B=b2 and C=c3.

Once a decision network has been defined, there are a variety of algorithms and software tools for processing the network (Buede, 1992). These algorithms can be used to derive the expected utility of any decision, or the posterior probability of any chance node, conditioned on specific values for any subset of the chance nodes.

### 2.1 DEFAULT TREES

We define a *default tree* (DTree) as a tree composed of default nodes (Dnodes) and evidence nodes (Enodes). Each Dnode specifies a decision, while each Enode specifies both an evidence item and a decision. To illustrate, the DTree in Figure 2 contains 4 Enodes and 6 Dnodes. This DTree corresponds to a decision procedure where the decision maker begins by either selecting d1 or examining evidence item A. If A is examined and its value is a2, then the decision d1 is immediately selected. If A=a1, then the decision maker selects d2 or examines B. If B=b1, then the decision d1 is selected, else if B=b2 then d2 is selected. Returning to the root Enode, if A=a3 then d3 is selected or C is examined. If C=c1, then d1 is selected. If C=c2 then d3 is selected or B is examined. If B=b1, then d2 is selected, otherwise d3 is selected.

Unless otherwise noted, we will assume that processing of a DTree continues until a Dnode is reached. That is, the evidence item associated with an Enode is always examined and the decision associated with an Enode is never selected.

Dnodes are partitioned into two types. A Dnode is *closed* if the path leading to the Dnode contains all the evidence items available. A Dnode is *open*, if it is not closed. Note that an open Dnode represents a *default decision*, since it specifies decisions that could change if additional evidence is examined.

More formally, we can characterize a DTree as follows. Let DN be a decision network which contains decision nodes {$D_j$} and evidence nodes {$E_i$}. Let DT be a DTree that contains the nodes {$N_i$}. Each member of {$N_i$} is an open Dnode, a closed Dnode, or an Enode. The following functions are defined with respect to DN.

$path_{DT}(N_i)$ - The set of evidence values in the ancestors of $N_i$.

For example, if we order the node in Figure 2 left-to-right breadth-first then $N_5$ is the left-most Dnode and $path_{DT}(N_5) = \{A=a1, B=b1\}$.

$evid\text{-}path_{DT}(N_i)$ - The evidence items that are listed in the ancestors of $N_i$ (e.g., $evid\text{-}path_{DT}(N_5)=\{A,B\}$). For the root node, $evid\text{-}path_{DT} = \{\}$.

$dec_{DT}(path_{DT}(N_i))$ - The maximum expected utility decisions in DN given the evidence item values leading to $N_i$. That is, $dec_{DT}(path_{DT}(N_i)) = \max_{dc}[EU(dc|path(N_i)]$, where dc is a set that specifies all possible combinations of decisions for the decision nodes in DN (e.g., $dec_{DT}(path_{DT}(N_5))=\{d1\}$).

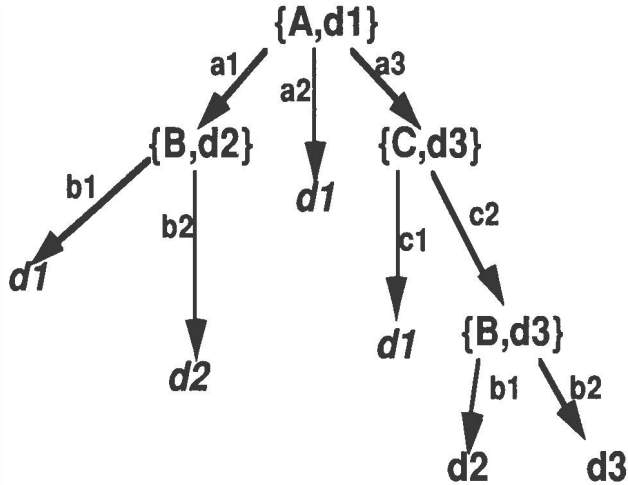

Figure 2. A Default Tree (DTree).

Let d be an open Dnode in the DTree DT and E an evidence item. We say that the *expansion* of d with E in DT is the DTree that results from replacing the Dnode d with the Enode {E,d}, and adding Dnodes for each possible value of E. Each new Dnode contains the maximum expected utility decisions. If Es is a sequence of evidence values, followed by an evidence item, then the *expansion sequence* of d with Es in DT is the DTree that results by starting at d and sequentially expanding DT with the evidence items in Es. For instance, the expansion of d with Es={E1=e1, E2=e2, E3} is obtained by replacing d with the expansion of d with E1, then replacing the Dnode at E1=e1 with the expansions of that Dnode with E2, and then replacing the Dnode at E2=e2 with the expansion of that node with E3. An *expansion subtree* is composed of one of more expansion sequences that have the effect of adding a subtree to the DTree. An *expansion set* is composed of multiple expansion subtrees.

$evoi_{DT}(E|path(N_i))$ = The increase in expected utility for examining evidence item E given the path to $N_i$. Formally,
$evoi_{DT}(E|path(N_i))$ =
$(\sum_{e \in E} P(E=e|path_{DT}(N_i))$
$\cdot EU[dec_{DT}(\{E=e\} \cup path_{DT}(N_i))])$
$- EU[dec_{DT}(path_{DT}(N_i))]$.

$max\text{-}evoi_{DT}(N_i)$ - The evidence item, E, for which $evoi_{DT}(E|path_{DT}(N_i))$ is maximal.

$eu\text{-}expand_{DT}(N_i,Es)$ - The increase in the expected utility of DT that is obtained by replacing $dec_{DT}(path(N_i))$ with the expansion subtree Es in DT.

It is easily shown that if Es contains a single evidence node (E), then

$eu\text{-}expand_{DT}(N_i,E) = P(path_{DT}(N_i)) \cdot evoi_{DT}(E|path(N_i))$.

If Es is an expansion subtree, then define *mean-eu-expand$_{DT}$(Es)* to be the mean value of the individual *eu-expand$_{DT}$* values for the evidence items in Es.

*Definition* (DT-compile)
  A DTree (DT) DT-compiles a decision network (DN) iff every evidence state in DN will lead to a Dnode in DT.

The DTree in Figure 2 DT-compiles the decision network in Figure 1.

*Theorem 1.* (DTree Expected Utility).
  If a DTree (DT) DT-compiles a decision network DN, then the expected utility of the DT, with respect to DN, is

  $EU(dec_{DT}()) + \sum_{N \in Enodes(DT)} eu\text{-}expand_{DT}(N,e)$,

  where e is always set to the evidence item at N.

*Proof.* (By induction)
  Let $\{N_i\}$ be a list of the Enodes in DT ordered in a manner that is consistent with partial ordering induced by the arcs in DT. Let $\{N_j\}_m$ be a subset of the first m Enodes in $\{N_i\}$. Each $\{N_j\}_m$ corresponds to a DTree. By definition, $EU(\{N_j\}_1) = eu\text{-}expand(N1) + EU(dec())$. Assume that $EU(\{N_j\}_m) = \sum_{N \in \{N_j\}_m} eu\text{-}expand_{DT}(N)$. Again by the definition of *eu-expand* $EU(\{N_j\}_{m+1}) = EU(\{N_j\}_m) + eu\text{-}expand_{DT}(N_{m+1})$.

## 2.2 DERIVING DTREES

The following algorithm can be used to derive a sequence of increasingly complex DTrees.

**Algorithm DD**
I. Let $N_1$ be a Dnode containing only $dec_{DT}()$.
II. Iterate through the following procedure
  A. Select the Open Dnode and evidence item for which $eu\text{-}expand_{DT}$ is maximal. Call this node N.
  B. Set N equal to the Enode $\{max\text{-}evoi_{DT}(N), dec_{DT}(path_{DT}(N))\}$
  C. For each possible value (e) of $max\text{-}evoi_{DT}(N)$ add as a subnode to N the Dnode $dec_{DT}(path_{DT}(N)$ & $max\text{-}evoi_{DT}(N)=e)$.
  D. Check stopping criterion. If Stop, then exist with current DTree.
  E. Go to A.

In words, DD iteratively replaces a default decision with the evidence item that maximizes the increase in the expected utility of the DTree, and adds as new subnodes the Dnodes that correspond to the best decisions for each possible value of that evidence item.[1] This algorithm is consistent with the situation-action tree development algorithm described informally in Heckerman, et.al. (1989).

---

[1] Obviously, the efficiency of DD could be increased substantially by recording the results of the *evoi*

DD is a greedy algorithm. At each iteration, it expands the DTree by adding and expanding the node that has the greatest increase in the expected utility of the DTree. We refer to such expansions as *greedy expansions*. Although DD is a greedy algorithm, it has two useful optimality properties.

*Theorem 2*: (Local optimality of DD)
Each Enode added by DD is a locally optimal extension in that if DTx and DTy are both one step expansions of DT, then EU(DTx) ≥ EU(DTy).

*Proof.*
This follows immediately from Theorem 1 and the fact that DD always selects the maximum *eu-expand* expansion.

To further characterize the optimality properties of DD, the following definitions are offered.

*Definition*. (Optimal DTree)
DT* is an optimal DTree iff EU(DT*)≥EU(DT) for each DTree (DT) where DT has no more Enodes than DT*. Also, DT* is an *optimal expansion* of DT iff DT* is an expansion of DT and DT* is an optimal DTree.

*Definition*. (E-descending)
DT is an E-descending DTree iff for every open Dnode (D), and evidence item (E),

P(*path*(D))*evoi*$_{DT}$(E|*path*(D))
 ≥ P(*path*(D)∪ {$e_i$})*evoi*$_{DT}$(E|*path*(D ∪ {$e_i$})),
where {$e_i$} is any set of evidence values.

In words, a DTree is E-descending iff it is impossible to increase the *eu-expand* value of an evidence item by making it the last element in an expansion sequence.

*Property 3*: (Optimal Dnode selection).
Let DT and DT* be any DTrees where DT is E-descending, DT* is an expansion of DT, and DT* is optimal. If {d} is the set of Dnodes in DT with the maximum *eu-expand* value, then DT* contains an expansion of a node in {d}.

*Proof*
P1. Let d be a Dnode in DT and E1, ..., En the evidence items available at d ordered by their *evoi* value. From this ordering, it follows that *eu-expand*(E1|path(d))≥*eu-expand*(Ei|path(d)) for i≥2. From E-descending it follows that *eu-expand*(Ei|path(d))≥*eu-expand*(Ei|path(d)∪{$e_i$})) for all {ei}. From transitivity it follows that *eu-expand*(E1|path(d)) is greater than the *eu-expand* value of an E-node in any subtree rooted at d.

P2. Let d' be a Dnode in the set of Dnodes with the maximal *eu-expand* value. From P1 and transitivity, it follows that *eu-expand*(E1|path(d')) is greater than or equal to the *eu-expand* value of any E-node in any subtree rooted at an open Dnode in DT.

P3. Let DT* be any expansion of DT that does not contain an expansion of a Dnode in {d}. From P2 it follows that all Enodes in DT', that are not in DT, have an *eu-expand* value that is strictly less than *eu-expand*(E1|path(d')). Consequently, a DTree which is the same as DT* except that it replaces a terminal Enode with the *max-evoi* expansion of d' will have a greater expected utility than DT*. Consequently, DT* cannot be an optimal DTree.

P4. The contrapositive of P4 is that if DT* is optimal, then it contains an expansion of a node in {d}.

*Theorem 4*. (Optimal Dnode selection by DD)
If each DTree (DT) generated by DD is E-descending, then each Dnode selected by algorithm DD for expansion must be expanded in any optimal expansion of DT that contains nodes other than those in DT or {d}, where {d} is the set of Dnodes in DT with the maximum *eu-expand* value.

*Proof*
DD selects the Dnode with the maximum *eu-expand* value. Consequently, by Property 3 an expansion of that Dnode must be included in an optimal expansion.

E-descending is not a very stringent constraint. While it is possible to increase the *evoi* value of an evidence item (E) by inserting a path of evidence values ({$e_i$}) as ancestors to E, a violation of E-descending requires that

$$evoi_{DT}(E|path(D \cup \{e_i\})) > \frac{evoi_{DT}(E|path_{DT}(D))}{P(\{e_i\}|path_{DT}(D))}.$$

That is, the *evoi*$_{DT}$ value must increase by a multiplier of more than 1/P({ei}|*path*$_{DT}$(D)). This can only occur if the increase in the *evoi* value is substantial or if P({$e_i$}) is near one. Both of these are unlikely if {ei} involves more than one evidence item. Consequently, violations of E-descending will be infrequent and most violations that do occur will only involve two level expansions.

Regarding globally optimality, Theorem 4 states that DD always expands the Dnode that must be expanded in an optimal expansion, while Theorem 2 asserts that DD always performs a locally optimal expansion of that Dnode. Intuitively, these two properties suggest that E-descending is sufficient to guarantee that DD-generated DTrees are optimal. It isn't. This is because greedy expansions may be redundant given several follow-on expansions. For instance, the *eu-expand* of E1 may be greater than either E2 or E3, even though *eu-expand*(E1)=0 if E2 and E3 are already included. Consequently, an

---

calculations and keeping track of open and closed Dnodes. However, efficiency improvement will not change the order in which the DTree is expanded. Consequently, they are not presented here.

optimal multi-step expansion may include E2 and E3, but not E1.

Unfortunately, the conditions required to guarantee global optimality are very stringent. In effect, it is necessary to assume conditions that imply that for any N, any DTree consisting of N greedy expansions is optimal. It is easy to construct violations of this property where the violations first appear at arbitrary expansion depths. Furthermore, since all DTree expansion procedures eventually lead to the same fully-expanded DTree, all expansion procedures will eventually converge to the same value. As they converge on this common value, there is no reason to believe that the greedy procedures will consistently generate optimal DTrees. Consequently, short of exhaustively searching the space of possible DTrees, there does not seem to be a way to guarantee generation of optimal DTrees.

On the other hand, the fact that violations of E-descending are not likely to involve insertions of long evidence chains suggests that DD can be enhanced by examining expansions a more than one level deep.

**Algorithm $DD_n$**
I. Let N1 be a Dnode containing only $dec_{DT}()$.
II. Iterate through the following procedure
  A. For each Open Dnode find the expansion subtree of depth n or less for which the *mean-eu-expand* value is maximal.
  B. Select the Open Dnode for which the *mean-eu-expand* value found in A is maximal. Call this node N and its expansion subtree Es.
  C. Replace N with the subtree that resulted in the maximal *mean-eu-expand* $_{DT}(N)$.
  D. Check stopping criterion. If Stop, then exist with current DTree.
  E. Go to A.

DD1 is the same as DD. $DD_n$ is similar to DD, except that it will look n levels deep to find the expansion with the greatest average contribution to the expected utility of the DTree. We call such expansions *greedy n-step expansions*. Since $DD_n$ examines strictly more nodes than DD, it will usually generate DTrees with expected utility greater or equal to the DTrees generated by DD. However, since neither algorithm is globally optimal, this cannot be guaranteed.

$DD_n$ satisfies local optimality under weaker conditions than DD.

*Definition.* ($E_n$-descending)
  DT is an $E_n$-descending DTree iff for every open Dnode (D), and evidence item (E),
  $P(path_{DT}(D))evoi_{DT}(E|path(D))$
  $\geq P(path_{DT}(D) \cup \{e_i\})evoi_{DT}(E|path_{DT}(D \cup \{e_i\}))$
  where $\{e_i\}$ is any set of evidence values with cardinality not less than n.

Note, E-descending is equivalent to $E_1$-descending.

*Theorem 5*  (Local optimality of $DD_n$).
  If each DTree generated by $DD_n$ is $E_n$-descending, then the *mean-eu-expand* value of each expansion subtree selected by $DD_n$ is greater than or equal to the *mean-eu-expand* value of any alternative expansion set.

*Proof.*
  As shorthand, let *meu* = *mean-eu-expand*.
  P1. We first show that for any DTree, there is a maximal *meu* expansion set which is a subtree. Let M be a maximum *meu* expansion set of DT. If M is not a single subtree, then M must be composed of a set of subtrees $\{M1,...,Mk\}$, each of which has its root at an open Dnode in DT.. Select a subtree Mi in $\{M1,...,Mk\}$ for which $meu(Mi) \geq max[meu(M1),...,meu(Mk)]$. From basic algebra it follows that $meu(Mi) \geq meu(M)$.
  P2. Next we show that for any open Dnode there is an expansion subtree with depth no greater than n for which *meu* is maximal. (The root node of an expansion subtree is at depth 1.) Let M be a maximal *meu* expansion subtree that contains at least one node with depth greater than n. Let $\{m1,m2,...,mi,mi+1,...,mk\}$ be the Enodes in M, where m1 is the root, m2,..,mi are all the nodes of depth n or less. Since M is a maximal *meu* expansion subtree, it follows that $meu(M) = meu[\{m1,m2,...,mi,mi+1,...,mk\}] \geq meu[\{m1,m2,...,mi\}]$; otherwise mi+1,...,mk would not be included in M. From basic algebra it follows that $meu[\{mi+1,...,mk\}] \geq meu[M]$. Let E* be the evidence item with the maximal *eu-expand* value. From $E_n$-descending it follows the *eu-expand*(E*) is greater than the *eu-expand* value of any possible node at depth n or greater. Therefore, $eu-expand(E*) \geq max(eu-expand((mi+1),...,eu-expand(mk))$, which implies $meu(E*) \geq meu(mi+1,...,mk)$. Therefore, $meu(E*) \geq meu(M)$. Consequently, the expansion subtree $\{E*\}$ is also a maximum value expansion subtree.
  P3. DDn always selects the expansion subtree with depth $\leq$ n with a greatest *meu* value. Therefore, it follows from P1 and P2 that DDn always selects an expansion set for which *meu* is maximal.

Algorithm $DD_n$ allows the decision network compilation process to be arbitrarily conservative. Indeed, if n is set to the number of evidence items, then $DD_n$ will exhaustively search the space of all DTrees. However, as noted above, violations of E-descending that are greater than two levels deep are very unlikely. Consequently, algorithm DD4 will almost certainly generate a sequence of expansions that are locally optimal for *any* search depth. $DD_n$ also satisfies the optimal Dnode selection property described in Property 4. This is because, whenever a DTree is E-descending, $DD_n$ will select the same expansion as DD.

## 2.3 COMPUTATIONAL COMPLEXITY

Let CI be the average computational complexity of processing the decision network. Let NE be the average number of evidence values for each evidence item. With each iteration of algorithm $DD_n$, the number of times the decision network is processed in $NE^n$. If a DTree contains R nodes, then the computational complexity of generating that DTree was $NE^n(R)(CI)$. That is, if the size of the DTree if fixed *a priori*, the computational burden of generating a DTree using $DD_n$ is a linear function of the computational burden of processing the decision network.

## 3.0 APPLICATIONS

In many organizations, the behavior of agents within that organization can be characterized as rule-guided. This is because the behavior of that organization is guided by a series of policy and procedure rules. Consider, for instance, air traffic control systems. The behavior of a ground control team is guided in large measure by a set of procedural rules, which specify how the team should react to various circumstances. The procedure rules specify conditions for de-icing, rerouting, priorities for landing, etc. Policy rules, in turn, provide guidelines for the establishment of the procedure rules (e.g., In snow, aircraft should be de-iced no less than one half hour before takeoff.)

A proposed set of rules for governing an organization's behavior can be evaluated in several ways. One way is to evaluate them in terms of their logical *consistency* and *completeness*. Do the rules always result in a consistent recommendation, or can different rule subsets lead to different actions? Do they specify what to do under all circumstances? Alternatively, rules can be evaluated in terms of their *executability*. Although a rule set may be internally consistent, it may be difficult define an acceptable architecture that can execute those rules (e.g., an architecture with a small number of communication links.) Finally, one can look at the expected *performance* of a proposed rule set. Performance evaluation presumes a model of the decision situations that a rule set is designed to handle, along with assessments of the probabilities and utilities associated with those situations. Otherwise, it would be possible to make a rule set look arbitrarily good or bad by carefully selecting the decision situations the rule set is tested against.

Each form of evaluation can provide a guide to the process of *generating* rule sets. For instance, Remy and Levis (1988) and Zaidi (1991) use concepts of architectural acceptability to derive a space of candidate architectures. These architectures, in turn, limit the types of rule sets that can be generated. The principal result of this paper is that a performance evaluation model can be used to derive procedure rules. In particular, the probability/utility information that is needed to evaluate a rule set is "compiled" into a DTree which defines a rule set that is logically complete, consistent, humanly-executable, and near-optimal in expected utility.

Note also, that the DTree formulation supports adaptation to temporal and workload constraints. Recall that there is a default decision associated with each node in a DTree. As a result, processing of a DTree can be terminated at anytime with a decision. This behavior can be represented within the DTree formulation by inserting additional Enodes, where time/workload information is the evidence that is examined. If there are severe time/workload constraints, then the Enode branches to a Dnode with a default decision. If processing time is available, then the Enode branches to the next evidence to consider.

## 4.0 FUTURE WORK

Future work in this area will address a number of important issues. The first is an empirical question. What is the expected size of a DTree? As noted earlier, a DTree is intended to be near-optimal. Furthermore, the expected utility of a DTree increases with each Enode that is added, with an asymptotic value equal to the expected utility of the decision network. However, it is remains an open question as to how large a near optimal DTrees must be. Second, there is the problem of time-varying value of information. The current formulation does not examine evolving situations, where the value of an item of information may change over time. Modified algorithms to consider time dependencies are being examined. Third, there is the problem of adaptive decision making. Although the DTree formulation effectively supports adaptation to time stress, it does not effectively support adaptation to other types of problems (e.g., sensor failures).

Finally, we note that the overall objective of this research is to develop near-optimal decision procedures that can be quickly and reliably executed by a team of human decision makers. The specification of a DTree is the first step in the process of specifying a team's decision procedures. The DTree must still be partitioned into several decision procedures that can be allocated to different team members. Work in this area is also proceeding.

### Acknowledgement

This research was supported by the Office of Naval Research under contract No. N00024-90-J-1680.

### References

Buede, D.M. (1992) Superior design features of decision analytic software, *Computers and Operations Research*, 19(1), 43-58.

Cooper, G.F. (1990) The computational complexity of probabilistic inference using Bayesian belief networks, *Artificial Intelligence*, 42, 393-405.

Dean, T.L. and Wellman, M.P. (1991), *Planning and Control*. Morgan Kaufmann, San Mateo, California.

Heckerman, D., Breese, J. and Horvitz, E. (1989) The compilation of decision models, in *Proceedings of the*

*1989 Workshop on Uncertainty in Artificial Intelligence*, 162-173.

Henrion, M. (1988) Propagating uncertainty in Bayesian networks by probabilistic logic sampling, in *Uncertainty in Artificial Intelligence 2*, J.F. Lemmer and L.N. Kanal (eds.), North Holland, Amsterdam.

Paass, G. (1991) Integrating probabilistic rules into neural networks: A stochastic EM learning algorithm, *Proceedings of the Seventh Conference on Uncertainty in Artificial Intelligence*, 264-270.

Pearl, J. (1987) *Probabilistic Reasoning in Intelligent Systems: Networks of Plausible Inference*. Morgan Kaufmann, San Mateo, California.

Remy, P. and Levis, A. H. (1988). "On the Generation of Organizational Architectures Using Petri Nets." in *Advances in Petri Nets 1988, Lecture Notes in Computer Science*, G.Rozenberg Ed. Springer-Verlag, Berlin, Germany.

Zaidi, S. A. K., "On the generation of Multilevel, Distributed Intelligence Systems using Petri Nets," MS Thesis, Report No.: GMU/C3I-113-TH, C3I Center, George Mason University, Fairfax, VA. November 1991.